\documentclass[twoside]{article}
\usepackage[accepted]{edge_dnetworks}

\usepackage[utf8]{inputenc} 
\usepackage[T1]{fontenc}    
\usepackage{hyperref}       
\usepackage{url}            
\usepackage{booktabs}       
\usepackage{amsfonts}       
\usepackage{nicefrac}       
\usepackage{microtype}      
\usepackage{algorithm2e}
\usepackage{algorithmic}
\usepackage{amsmath,amsfonts,amssymb}       
\usepackage{bm}     
\usepackage{graphicx}     
\usepackage{caption}       
\usepackage{hhline}       

\newcommand{\Gt}{G^{(t)}}
\newcommand{\Vt}{V^{(t)}}
\newcommand{\Et}{E^{(t)}}
\newcommand{\eit}{e^{(t)}_{i}}
\newcommand{\cit}{c^{(t)}_{i}}
\newcommand{\vit}{v^{(t)}_{i}}
\newcommand{\vpit}{v^{{\prime}^{(t)}}_{i}}

\newcommand{\kt}{\mathbf{k}^{(t)}}

\newcommand\numberthis{\addtocounter{equation}{1}\tag{\theequation}}
\graphicspath{{imgs/}}
\begin{document}

\twocolumn[

\aistatstitle{A Dynamic Edge Exchangeable Model for Sparse Temporal Networks}

\aistatsauthor{ Yin Cheng Ng \And Ricardo Silva}

\aistatsaddress{y.ng.12@ucl.ac.uk\\Statistical Science\\University College London \And ricardo.silva@ucl.ac.uk\\Statistical Science\\University College London} ]
\begin{abstract}
We propose a dynamic edge exchangeable network model that can capture sparse connections observed in real temporal networks, in contrast to existing models which are dense. The model achieved superior link prediction accuracy on multiple data sets when compared to a dynamic variant of the blockmodel, and is able to extract interpretable time-varying community structures from the data. In addition to sparsity, the model accounts for the effect of social influence on vertices' future behaviours. Compared to the dynamic blockmodels, our model has a smaller latent space. The compact latent space requires a smaller number of parameters to be estimated in variational inference and results in a computationally friendly inference algorithm.
\end{abstract}

\section{Sparse Temporal Networks}\label{sec:temporal}
We consider a temporal sequence of networks $\{\Gt\}_{t=1}^T$ indexed by $t\in \mathbb{Z}^+$, each containing a set of vertices and edges $\Gt\equiv{\{\Vt, \Et\}}$. $\Et = \{e^{(t)}_{1}, ..., e^{(t)}_{N^{(t)}}\}$ is the set of $N^{(t)}$ edges observed at $t$ and $\Vt$ is the set of vertices that have participated in at least one edge up to $t$ such that $V^{(t-1)}\subseteq\Vt$. An edge $\eit = (\vit, \vpit)$ is a tuple of two interacting vertices, and may or may not be directed. We focus on undirected temporal networks in this paper and ignore the order of vertices in the tuple. Many events that involve pairwise interactions between entities and individuals can be considered as temporal networks. Real-world examples of temporal networks include e-mail communication networks, friendship networks, trading networks and many more. 

Temporal networks exhibit statistical properties that are of practical interest. We focus on addressing three important properties in this paper: sparsity, community structure and social influence. Our proposed model differs from the existing models by taking into account all three properties simultaneously while being less computationally demanding compared to many existing ones. It also allows the set of vertices to grow over time, as opposed to forcing the set of vertices to remain constant.

We discuss the three properties in the remaining sub-sections, and introduce the proposed model in Section~\ref{sec:proposed}. We then review some existing models for temporal networks and compare them to the proposed model in Section~\ref{sec:existing}. In addition, we discuss some related works in Section~\ref{sec:related}. We close the paper with some experimental results and discussions in Section~\ref{sec:exp}.

\subsection{Sparsity}
The connections observed among the vertices in real-world networks are typically sparse, with only a small number of observed edges compared to all possible pairs of vertices \cite{barabasi2016network,goldenberg2010survey}. Using a social network with thousands of members as an intuitive example, most, if not all members of the social network are connected to tens or hundreds of other members in the network, instead of thousands of other members. Therefore, the total number of connections in the social network is an order of magnitude smaller than the number of all possible pairs of members. Sparsity is an essential condition to maintain certain structural properties observed in real networks, such as the `small world phenomena' and the power-law degree distributions, as these properties can only occur in sparse networks \cite{orbanz2015bayesian}. Formally, a network $G = (V,E)$ with $|E|$ edges and $|V|$ vertices is sparse if $|E| = o(|V|^2)$ (i.e., $|E|$ is asymptotically upper bounded by $c \cdot |V|^2$ for $c>0$).

In the context of temporal networks $\{\Gt\}_{t=1}^T$, each of the observed networks $\Gt$ may be sparse. We argue that in some cases, $\Gt$ is sparser than static networks that are aggregates of temporal networks. Therefore, it is important that temporal network models are able to capture the sparse property of real observations. However, as we discuss in Section~\ref{sec:existing}, most existing dynamic network models do not allow for sparse connections because of their underlying exchangeability assumption.

\subsection{Community Structure}
In many networks, the connections between vertices can be clustered into different categories, with overlapping subsets of vertices dominating each of the categories. In a social network, for example, edges can represent relationships between colleagues, college friends, family members and other types of social relationships. Within each type of relationship, a subset of vertices are over-represented compared to the others. Some vertices may also dominate multiple types of relationships. Vertices that participate in multiple types of connections are referred to as having mixed-memberships in the stochastic blockmodel literature \cite{airoldi2008mixed}. In temporal networks, the number of edges belonging to each category can fluctuate through time and the dominating vertices of each category can also evolve \cite{xing2010state,xu2014dynamic,ho2011evolving}. The types of edges are not annotated in many network data sets, but can be inferred through statistical analysis. While dynamic variants of the mixed-membership stochastic blockmodel can model the mixed-membership community structure, they cannot account for sparse connections by construction \cite{orbanz2015bayesian}. As a consequence, these models do not allow for the presence of hubs and power-law degree distributions.

\subsection{Social Influence}
The presence of an edge connecting two vertices at time $t$ implies that the two vertices had interacted in some ways during the period. The interaction may influence the states of both vertices at the subsequent time point, causing the two vertices to interact with the world similarly in the future. The two vertices may also interact with many other vertices at $t$, causing their respective future states to reflect their own historical states and the social influence from their respective neighbours at $t$ to various degrees. The causal nature of time allows us to draw potential causal links from the observed connections of vertices at $t$ to their future states, and perform inference on future connections between vertices \cite{blundell2012modelling, farajtabar2015coevolve, linderman2014discovering}. The effect of social influence plays an important role in the evolution of temporal networks, and should be taken into account in building temporal network models \cite{heaukulani2013dynamic}.

\section{Dynamic Edge Exchangeable Network Model}\label{sec:proposed}
We propose a dynamic model for sparse temporal networks $\{\Gt\}_{t=1}^T$ that is built upon the edge exchangeable framework proposed in \cite{cai2016edge,crane2016edge}. By enforcing the edge exchangeable assumption to the marginal distribution at each time point $t$, the model allows $\Gt$ to be sparse. In contrast, the existing dynamic models guarantee that $\Gt$ is either empty or dense (see Section~\ref{sec:existing}). The edge exchangeable marginals for different time points are coupled by latent Gaussian Markov chains to model the social influence effects and evolution of the temporal networks. Additionally, we introduce a Poisson vertex birth mechanism to allow new vertices to join the networks at different times. In the following sub-sections, we first discuss the generative process for individual $\Gt$ and introduce the Markov dynamics that couple together the temporal sequence of networks. We then present a variational inference algorithm and discuss its computational complexity in Section~\ref{sec:vi}.

\subsection{Edge Exchangeable Sparse Networks}
The edge exchangeable assumption applied to each of the networks $\Gt$ in the temporal sequence $\{\Gt\}_{t=1}^T$ dictates that the edges in the set $\Et = \{e^{(t)}_{1}, ..., e^{(t)}_{N^{(t)}}\}$ are exchangeable and the probability distribution of $\Gt$ is invariant to the order of the edges in $\Et$. Therefore, $e^{(t)}_{1}, ..., e^{(t)}_{N^{(t)}}$ are $i.i.d.$ samples of an edge distribution $P_t(e)$. Following our notations from Section~\ref{sec:temporal}, an undirected edge $\eit$ is a tuple of two unordered participating vertices $(\vit, \vpit)$. As a result, $P_t(e = (\vit, \vpit))$ can be factorized into a product of identical vertex distributions $P_t(\vit)P_t(\vpit)$. We present some simulation results in Section~\ref{sec:exp} to demonstrate that our edge exchangeable network construction can model sparsity. We refer the readers to \cite{janson2017edge} for a detailed general review of edge exchangeable networks.

\subsection{Community Structure Mixture Model}
To incorporate community structure, we adopted the edge clustering approach proposed for non-parametric static network model in \cite{williamson2016nonparametric} to our model. We assume that the observed edges $\Et$ are samples of a mixture of $M$ edge distributions in Equation~\ref{eqn:mixture} (corresponding to $M$ communities) where the per-edge latent mixture component indicators $\cit \in \{1, ..., M\}$ describe the types of connection that the edges belong to. We suppress the parameters in Equation~\ref{eqn:mixture} for compact presentation.
\begin{align*}\label{eqn:mixture}
    P_t&(e=(\vit,\vpit)) = \sum_{m=1}^M[P_t(\cit=m)\numberthis\\
    &P_t(\vit|\cit=m)P_t(\vpit|\cit=m)]
\end{align*}
For undirected networks, $P_t(\vit|\cit)$ and $P_t(\vpit|\cit)$ are identical vertex distributions which we collectively denote as $P_t(v|\cit)$ for notational convenience. $P_t(\cit)$ and $\{P_t(v|\cit=m)\}_{m=1}^M$ are parameterized as logistic normal distributions with $M$-dimensional and $(|V^{(t-1)}|+L^{(t)})$-dimensional support respectively. $|V^{(t-1)}|$ is the number of vertices observed in the networks up to the previous time point (with $|V^{(0)}|=0$) and $L^{(t)}$ is the Poisson distributed number of potential new vertices that may join the network at $t$. 

The relative sizes of the communities may grow and shrink over time and exhibit temporal dependency, as sudden large changes to the community sizes are rare. We construct the following latent Gaussian Markov chain for the parameters of the logistic normal distributions $\mathbf{k}^{(t)} \in \mathbb{R}^{M\times 1}$ to capture the temporal dependency.
\begin{equation}\label{eqn:cls_dist}
    P_t(\cit=m|\mathbf{k}^{(t)}) \propto e^{k_m^{(t)}}
\end{equation}
\begin{align*}\label{eqn:mixing}
    P(\mathbf{k}^{(1:T)}) = &\mathcal{N}(\mathbf{k}^{(1)};\bm{\mu}_k, \mathbf{B}_k\mathbf{B}_k^\intercal)\numberthis\\ 
    &\prod_{t=2}^{T}\mathcal{N}(\mathbf{k}^{(t)};\mathbf{A}_k\mathbf{k}^{(t-1)}, \mathbf{B}_k\mathbf{B}_k^\intercal)
\end{align*}
$\bm{\mu}_k \in \mathbb{R}^{M\times 1}$, $\mathbf{A}_k \in \mathbb{R}^{M \times M}$ and the lower-triangular matrix $\mathbf{B}_k \in \mathbb{R}^{M \times M}$ are the parameters of the Markov chain.

The $m^{th}$ vertex distribution at $t$, $P_t(v|\cit=m)$ (abbreviated as $P_{t,m}(v)$ for compactness), encodes the relative dominance of the vertices in community $m$ at time $t$. We endow each vertex $v$ at time $t$ with a Gaussian distributed latent state vector $\mathbf{h}_v^{(t)} \in \mathbb{R}^{M \times 1}$, such that 
\begin{equation}\label{eqn:vertex_emit}
    p_{t,m}(v=v_i) \propto e^{h_{v,m}^{(t)}}.
\end{equation}
The normalizing constant for Equation~\ref{eqn:vertex_emit} is the sum over the exponentiated $m^{th}$ hidden states of all the vertices in $V^{(t-1)}$ and the $L^{(t)}$ potential new vertices at $t$.

\subsection{Markov Dynamics with Social Infuence}\label{sec:markov}
The latent state vectors of the $L^{(t)}$ potential new vertices are sampled from an initial Gaussian distribution parameterized by the mean vector $\bm{\mu} \in \mathbb{R}^{M\times 1}$ and the lower-triangular matrix $\mathbf{B} \in \mathbb{R}^{M \times M}$
\begin{equation}\label{eqn:vertex_init}
    p(\mathbf{h}) = \mathcal{N}(\mathbf{h};\bm{\mu}, \mathbf{B}\mathbf{B}^\intercal).
\end{equation}
If the potential new vertices are indeed sampled to form $\Gt$, they are added to the set $\Vt$ and together with the existing vertices that joined in the previous time steps, their respective latent state vector $\mathbf{h}_v^{(t)}$ evolves to the next time point according to the conditional Gaussian distribution in Equation~\ref{eqn:vertex_evolve}. The new potential vertices at $t$ that do not participate in $\Gt$ are dropped from the model and ignored in the next time step.
\begin{align*}\label{eqn:vertex_evolve}
    p(&\mathbf{h}_v^{(t+1)}|\Gt, \{\mathbf{h}_i^{(t)}|i\in\Vt\}) \numberthis\\
    &= \mathcal{N}(\mathbf{h}_v^{(t+1)}; \mathbf{f}(v, \Gt, \{\mathbf{h}_i^{(t)}|i\in\Vt\}), \mathbf{B}\mathbf{B}^\intercal)
\end{align*}

To model the varying degrees of social influence on the evolution of temporal networks, the mean of the conditional Gaussian in Equation~\ref{eqn:vertex_evolve} is parameterized as a $M$-dimensional vector function $\mathbf{f}_{v,t} = \mathbf{f}(v, \Gt, \{\mathbf{h}_i^{(t)}|i\in\Vt\})$ defined in Equation~\ref{eqn:attn}.
\begin{align*}\label{eqn:attn}
    \mathbf{f}_{v,t} = w_{vv}^{(t)}\mathbf{h}_v^{(t)} + \sum_{i\in ne(v,t)}w_{vi}^{(t)}\mathbf{h}_i^{(t)} \numberthis
\end{align*}
where $ne(v,t)$ is the set of neighbour vertices that vertex $v$ formed an edge with in $\Gt$. Equation~\ref{eqn:attn} is a weighted average of $\mathbf{h}_v^{(t)}$ and the previous latent state vectors of vertex $v$'s neighbours at $t$. The non-negative weights $w_{vi}^{(t)}$ is a dot-product based similarity measure between vertex $v$ and $i$ defined as 
\begin{align*}\label{eqn:weights}
    w_{vi}^{(t)} = \frac{e^{\mathbf{h}_v^{(t)} \cdot \mathbf{h}_i^{(t)}}}{e^{\mathbf{h}_v^{(t)} \cdot \mathbf{h}_v^{(t)}}+\sum_{j \in ne(v,t)}e^{\mathbf{h}_v^{(t)} \cdot \mathbf{h}_j^{(t)}}}. \numberthis
\end{align*}
Intuitively, the neighbours of the vertex $v$ at $t$ pull the latent state of $v$ towards themselves at different degrees after they interacted at $t$. The neighbours that are more similar to vertex $v$ in the state-space have higher influence on its future latent state. If the vertex $v$ did not interact with any vertex, then $\mathbf{f}_{v,t}=\mathbf{h}_v^{(t)}$ and Equation~\ref{eqn:vertex_evolve} is simply a random walk. 

The weighted average parameterization of the conditional mean in Equation~\ref{eqn:attn} is similar to the local context-based soft attention mechanism proposed in \cite{luong2015effective} for NLP neural networks in two ways. Firstly, it assigns higher weights to vertices that are more similar in the latent space. Similarly, the attention mechanism in \cite{luong2015effective} assigns higher weights to words that are similar in context. Secondly, Equation~\ref{eqn:attn} avoids the computationally expensive operation of summing over all existing vertices by looking only at the neighbours of vertex $v$ at the previous time step. It is `local' in the sense that it only sums over vertex $v$'s immediate neighbours. The attention mechanism in \cite{luong2015effective} is `local' in the sense that the attention mechanism only consider other words that surround the target word in a sentence instead of the whole corpus, leading to saving in computational costs. We refer to the state-space parameterized by Equation~\ref{eqn:attn} as the Attention Augmented State-space (\textbf{ATTAS}). To the best of our knowledge, we are the first to propose an attention mechanism in probabilistic model for network data. We compare ATTAS to a simple random walk (\textbf{RW}) state-space as well as other models in the prediction experiments.

\subsection{Poisson Birth Mechanism}
The number of new vertices $L^{(t)}$ at time $t$ is uncertain prior to observing $\Gt$. We seek to account for the uncertainty with a Poisson prior distribution with log-rate parameter $\lambda^{(t)}$
\begin{equation}\label{eqn:poisson}
    P(L^{(t)}|\lambda^{(t)}) = Poisson(e^{\lambda^{(t)}}).
\end{equation}
We observed in many temporal networks that $L^{(1)}$ is typically large because no vertex existed in the networks prior to $t=1$. The subsequent $L^{(t)}$ typically become smaller. Therefore, we propose to capture the temporal dynamics of $L^{(t)}$ with an auto-regressive Markov chain prior on the parameter sequence $\lambda^{(1:T)}$
\begin{align*}\label{eqn:lambda}
    p(\lambda^{(1:T)}) = \mathcal{N}&(\lambda^{(1)}; \mu_\lambda; \sigma_\lambda^2) \numberthis\\
    &\prod_{t=2}^T\mathcal{N}(\lambda^{(t)};a_\lambda\lambda^{(t-1)}, \sigma_\lambda^2).
\end{align*}
In the scenario when $0<a_\lambda<1$, the long-run expectation of $L^{(t)}$ is $1$ despite the larger initial expectation of $e^{\mu_\lambda}$. As previously described, the $L^{(t)}$ potential new vertices are assigned a latent state vector sampled from Equation~\ref{eqn:vertex_init} and are discarded from the model if they do not participate in $\Gt$.

\subsection{Model Summary}
\begin{figure}[!htbp]
    \centering
    \includegraphics[scale=0.25]{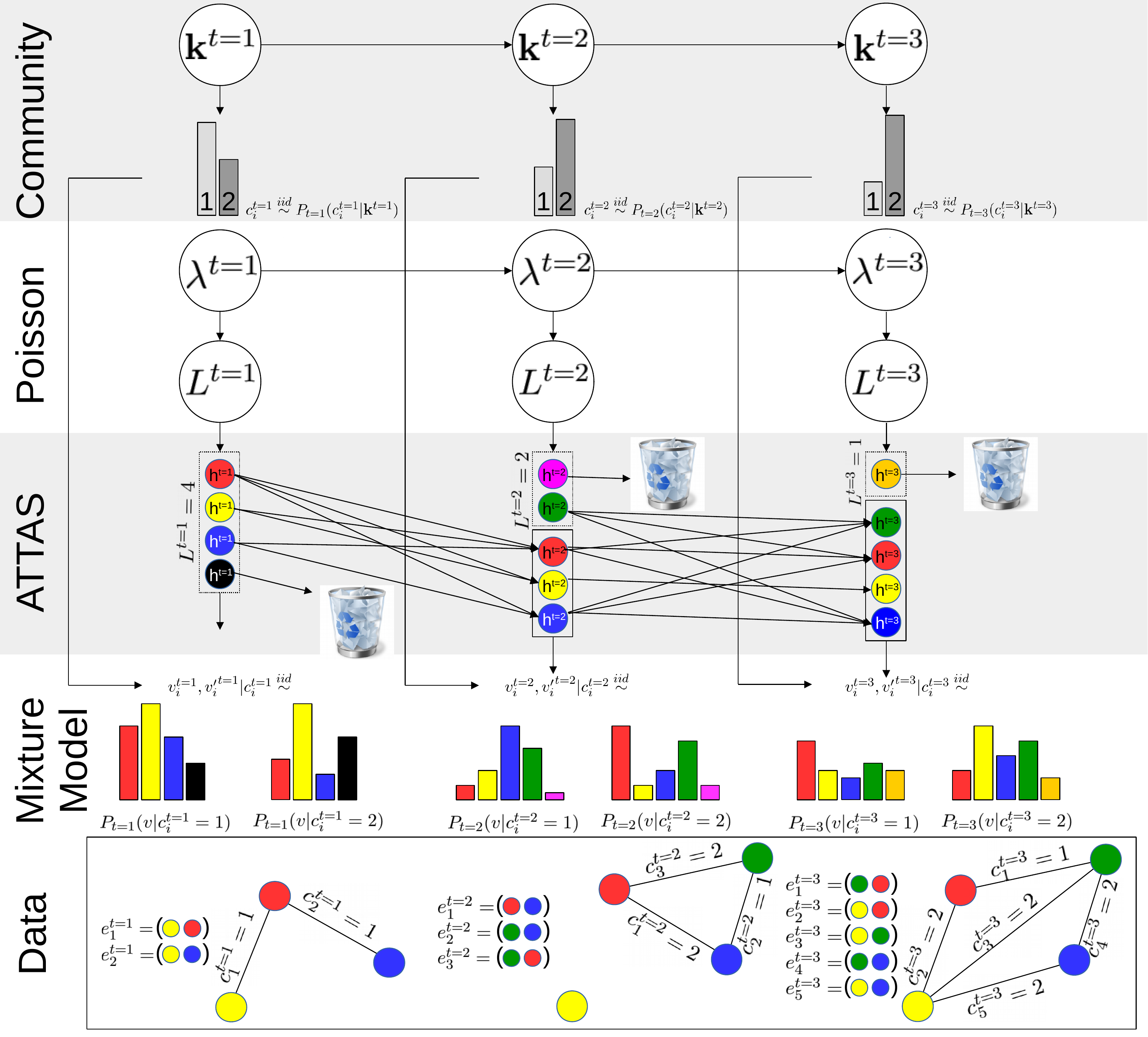}
    \caption{The figure shows the generative process for a sequence of 3 temporal networks with 2 communities. This figure is best understood together with the description and best viewed on a computer.}
    \label{fig:generative}
\end{figure}

We provide a generative summary of the dynamic edge-exchangeable network model described in the previous sub-sections, and depict the generative process for a sequence of $T=3$ temporal networks with 2 communities in Figure~\ref{fig:generative}. We assume that the number of edges sampled at each time point $N^{(1:T)}$ are directly specified. However, they can also be modeled with Poisson distributions or directly observed from the data sets in practice.

Given the number of communities $M$, $T$, $N^{(1:T)}$, and model parameters $\theta = \{\mu_\lambda$, $\sigma_\lambda$, $a_\lambda$, $\bm{\mu}$, $\mathbf{B}$, $\bm{\mu}_k$, $\mathbf{A}_k$, $\mathbf{B}_k\}$, we generate the temporal networks as follow.

For $t$ in $1, \ldots, T$: 
\small
\begin{enumerate} 
    \itemsep0em 
    \item Set $V^{(t)} = V^{(t-1)}$, $\Et = \{\}$ ($V^{(0)}\equiv \{\}$)
    \item Draw $\lambda^{(t)}, L^{(t)}, \kt$ from Equation~\ref{eqn:lambda}, \ref{eqn:poisson}, \ref{eqn:mixing}.
    \item For $i$ in $1, \ldots, L^{(t)}$, draw $\mathbf{h}_{z_i}^{(t)}$ from Equation~\ref{eqn:vertex_init}.
    \item For $v\in V^{(t)}$, draw $\mathbf{h}_v^{(t)}$ from Equation~\ref{eqn:vertex_evolve}.
    \item For $i$ in $1, \ldots, N_t$:
    \begin{enumerate}
    \itemsep0em
        \item Draw $\cit$ from Equation~\ref{eqn:cls_dist}.
        \item Draw $\eit=(\vit, \vpit)$ conditioning on $\cit$ from Equation~\ref{eqn:vertex_emit}.
        \item $\Et=\{\Et, \eit\}$; $\Vt=\Vt \cup \{\vit, \vpit\}$
    \end{enumerate}
\end{enumerate}\normalsize

\paragraph{Description of Figure~\ref{fig:generative}}
Individual vertices in Figure~\ref{fig:generative} are represented as colored balls in the data box, with pairs of vertices forming edges that compose the networks. The yellow vertex does not participate at $t=2$ but remains in the network because it participated at $t=1$ while the green vertex joins at $t=2$. The community labels of the edges are sampled from the grayscale community distributions at the top, and annotated on the sampled edges. Conditioning on the community label, two vertices are sampled from the colored mixture distributions above the box to form an edge. The vertex state vectors $\mathbf{h}_v^{(t)}$ of the mixture distributions are represented as colored circles in the rectangles above the distributions, with their ATTAS evolution mechanism described in Section~\ref{sec:markov} depicted as directed arrows through time. $\mathbf{h}_v^{(t)}$ of new potential vertices are grouped in dotted rectangles, with those that do not immediately participate in the network discarded at every time step (e.g., black at $t=1$). The number of new potential vertices $L^{(t)}$ is determined by Poisson distributions with log-rates $\lambda^{(t)}$ that evolve according to a Gaussian Markov chain.

\section{Variational Inference}\label{sec:vi}
We approximate the posterior distributions of the model's latent variables with a variational inference (VI) algorithm \cite{blei2017variational}. The latent variables of interest are $L^{(1:T)}$, $\lambda^{(1:T)}$, $\mathbf{k}^{(1:T)}$, the per-edge community type latent variables $\cit$ and the per-vertex latent state vectors $\mathbf{h}_v^{(\tau_{v}:T)}$. $\tau_{v}$ is the time when vertex $v$ first joined the networks.

The time dependency of temporal networks requires that the approximating variational distributions $q$ for $\mathbf{h}_v^{(\tau_{v}:T)}$, $\mathbf{k}^{(1:T)}$ and $\lambda^{(1:T)}$ to preserve their time dependency. To preserve the time dependency while allowing tractable variational distributions, we utilize the structured mean-field (SMF) family of variational distributions \cite{saul1996exploiting}. The SMF family approximates $q(\mathbf{h}_v^{(\tau_{v}:T)})$ for $v \in V^{(T)}$, $q(\mathbf{k}^{(1:T)})$ and $q(\lambda^{(1:T)})$ as Gaussian Markov chains \cite{blei2006dynamic, ghahramani1997factorial,ng2016scaling,archer2015black}, and the remaining variational distributions as fully-factorized mean-fields.

The proposed ATTAS state-space introduce non-conjugate structures to the model through the $\mathbf{f}_{v,t}$ function in Equation~\ref{eqn:attn}. Additionally, the log-normalizing constants of the multivariate logistic normal distributions in Equation~\ref{eqn:cls_dist} and Equation~\ref{eqn:vertex_emit} are also non-conjugate. The two sources of non-conjugacy render the evidence lower bound (ELBO) objective function of VI analytically intractable. We take a two-pronged approach to tackle the intractable ELBO. We first linearize the log-normalizing constants using Taylor's series approximation \cite{blei2006dynamic}. The linear approximation introduces an additional lower bound to the ELBO, but allows $\{\{q(\cit)\}_{i=1}^{N^{(t)}}\}_{t=1}^T$ to be optimized analytically through fast conjugate updates. The conjugate updates are equivalent to optimizing the variational parameters with natural gradients, and lead to faster convergence. To tackle the more complex intractable terms introduced by ATTAS, we resort to optimizing the variational parameters of $q(\mathbf{h}_v^{(\tau_{v}:T)})$ with ADAM \cite{kingma2014adam} using unbiased Monte Carlo gradients computed with the reparameterization tricks \cite{kingma2013auto,archer2015black}. We alternate between performing the conjugate updates and multiple steps of the stochastic gradient updates. We find that exploiting the linear approximations and conjugate updates lead to faster and better convergence when compared to a fully Monte Carlo approach.

We learn the model parameters $\theta$ by maximizing the same ELBO objective as VI. The model parameters updates are performed together with the stochastic gradient updates of VI.

We validated the goodness of the VI approximation using a simulated experiment to recover ground truth edge community labels $\cit$. The VI algorithm was able to recover $96\%$ of the 1694 ground truth labels across 3 time steps at a normalized mutual information (NMI) score of 0.75. The details of the simulation experiment, together with the VI algorithm derivations are available in the Supplementary Material.

\subsection{Computational Complexity}\label{sec:complexity}
The computational bottleneck of the variational inference algorithm lies in computing the $M \times T$ approximated expected log-normalizing constant terms and the corresponding gradients, contributed by the logistic normal distributions in Equation~\ref{eqn:vertex_emit}. Computing the approximated expectations requires summing over the expected exponentiated latent state vectors $\mathbf{h}_v^{(t)}$ for all vertices in $V^{(T)}$. The sums are then used to update $\{\{q(\cit)\}_{i=1}^{N^{(t)}}\}_{t=1}^T$. Therefore, the computational complexity of the VI algorithm is $O(E_{TOT}M + |V^{(T)}|MT)$, where $E_{TOT} = \sum_{t=1}^TN^{(t)}$, $|V^{(T)}|$ is the total number of vertices in the temporal networks, $M$ is the number of communities and $T$ is the number of time points. The computational complexity of VI for the proposed model is significantly lower than the dynamic variants of mixed-membership stochastic blockmodels, which have complexity of $O(M|V^{(T)}|^2T)$ and beyond \cite{xing2010state,ho2011evolving}.

\section{Comparisons to Existing Models}\label{sec:existing}
We compare and contrast the proposed dynamic edge exchangeable network model to some existing probabilistic models for temporal networks, focusing on how the proposed model handles the 3 key properties discussed in Section~\ref{sec:temporal} differently, and the models' inference computational complexities.

One key property of the proposed model is its ability to model sparse connections in each networks $\Gt$ in the temporal network sequence by assuming the edges observed within each time points are exchangeable units of data. This is in contrast to the existing models that are dynamic variants of the mixed-membership stochastic blockmodel \cite{xing2010state, ho2011evolving,xu2014dynamic,zreik2017dynamic}, latent space model \cite{sarkar2006dynamic,sewell2017latent} and latent feature model \cite{heaukulani2013dynamic}. Marginally, these existing dynamic models make use of likelihood models that fall under the exchangeable random graph framework of Aldous-Hoover representation theorem \cite{aldous1981representations,hoover1979relations}. It is well known that exchangeable random graphs are either empty or dense \cite{orbanz2015bayesian}. Therefore, under the existing models, $\Gt$ cannot be sparse. As we discussed in Section~\ref{sec:temporal}, certain structural properties of real networks are unique to sparse networks only. The limitation of the existing models in capturing sparsity is an important disadvantage.

The proposed model incorporates community structure by directly clustering edges with a mixture model, and interpret the per-edge latent mixture component indicators $\cit$ as the types of interactions between the two interacting vertices (e.g., work connections, college friends, family ties in social networks). The edge-clustering approach models the same type of community structure as the assortative mixed-membership stochastic blockmodel (aMMSB) \cite{gopalan2012scalable} despite the differences in model construction. In aMMSB, each of the vertices assumes different interaction-specific latent roles when interacting with other vertices and two vertices form an edge with high probability only when both vertices assume the same latent roles (e.g., colleague-colleague interactions etc.). The types of interactions encoded in aMMSB correspond to the interaction-specific latent roles of both interacting vertices, and is equivalent to the edge-clustering formulation. For example, a colleague-colleague interaction is equivalent to a work connection.

The direct edge-clustering approach of our model results in a significantly smaller set of latent variables compared to an equivalent aMMSB. To model a static network $G = (V, E)$ with $M$ communities, the proposed edge-clustering model requires $|E|$ latent mixture component indicator $\{c_i\}_{i=1}^{|E|}$ while aMMSB (or other types of MMSB) requires $2 \cdot |V|^2$ latent role indicators. Generalizing to temporal networks $\{\Gt\}_{t=1}^T$, the proposed model requires only $\sum_{t=1}^T|E^{(t)}|$ latent variables to capture the community structure compared to $2 \cdot T|V^{(T)}|^2$ for dynamic MMSBs. The difference in the size of latent space is especially significant when the networks $\{\Gt\}_{t=1}^T$ are sparse. The compactness of our model results in fewer numbers of variational distributions to approximate in variational inference, and lead to computational gains as discussed in Section~\ref{sec:complexity}.

The Dynamic Latent Feature Propagation model proposed in \cite{heaukulani2013dynamic} also accounts for the social influence of neighbours by assigning `social influence weights’ to the vertices. The model is restricted in that each vertex has equal influence on their neighbours. The ATTAS proposal bypassed the restriction with the attention mechanism, such that the vertices exert higher influence on their neighbours that are more similar. In addition, the ATTAS construction does not introduce any additional model parameters that need to be learned while the social influence weights of the Dynamic Latent Feature Propagation model require optimization.

\section{Related Work}\label{sec:related}
In addition to the work on dynamic networks mentioned previously, the ATTAS state-space proposed in Section~\ref{sec:proposed} also relates to continuous-time point processes models for reciprocating relationships such as \cite{blundell2012modelling,farajtabar2015coevolve,linderman2014discovering}. These models are typically applied to a fixed number of nodes and correspond to dense generative models, like MMSBs, which provide no simple control on the number of sampled interactions within any particular time window: each interaction follows directly from node-level or node-cluster-level latent features, as opposed to edge-cluster-level features. Continuous-time models have advantages and disadvantages compared to discrete-time models that are well-studied. We favour discrete-time modeling due to the inferential ease by which we can enforce a bottleneck of event counts, and the flexibility of attention models as way of parameterizing interactions using individual-level latent vectors.
 
There have been many advances on nonparametric sparse network models \cite{caron2014sparse,cai2016edge,crane2016edge,williamson2016nonparametric}, with efficient Markov chain Monte Carlo algorithms. However, existing works on relating these models to dynamic modeling is limited \cite{williamson2016nonparametric}. Models based on nonparametric block models exist in the literature (e.g., \cite{ishiguro2010dynamic}). It is not our goal to fill in the gap between continuous-time dynamic and nonparametric edge-clustering models. In practice, we believe that social data is too non-stationary for the elegant machinery of Hawkes processes to really be effective, and we see our contribution as a practical framework for short term predictions and historical smoothing and clustering of observed interactions over which extensions can be built.

\section{Experiments}\label{sec:exp}
We conducted 3 experiments with the following goals.
\begin{enumerate}
    \itemsep0em 
    \item Investigate sparsity under various hyper-parameter settings.
    \item Benchmark the model's link prediction powers.
    \item Investigate the model's capacity to capture community structures.
\end{enumerate}

\subsection{Sparse Networks Simulations}
The proposed dynamic network model consists of $T$ logistic normal edge distributions that are coupled by latent Markov chains. Each edge distribution can capture sparse connections in $\Gt$. In this experiment, we investigate the sparsity of networks simulated from a logistic normal edge distribution (i.e., $T=1$) with different hyper-parameter settings, and show that the networks simulated with certain hyper-parameter settings are sparse. To focus on sparsity, we set $M=1$ and ignore the community structure.

To simulate the networks, we first sample the number of latent vertices $L \sim Poisson(10^6)$ and the vertices' latent states $h_i \sim \mathcal{N}(0, \sigma^2)$ for $i \in \{1, \ldots, L\}$. The edges are then sampled from the edge distribution $P(e = (i, j)|\sigma) = P(v=i|\sigma)P(v=j|\sigma)$ where $P(v=i|\sigma) = \frac{e^{h_i}}{\sum_{l=1}^Le^{h_l}}$. The hyper-parameter of interest is the standard deviation $\sigma$. The standard deviation fully determines the excess kurtosis of the transformed random variable $e^{h_i}$, which in turn governs the shape of the logistic normal $P(v|\sigma)$. A small $\sigma$ value corresponds to a flat $P(v|\sigma)$ while a large $\sigma$ corresponds to a multi-modal $P(v|\sigma)$ because of the resulted heavy-tailed distribution for $e^{h_i}$. With a flat $P(v|\sigma)$, the number of vertices sampled to join the network (i.e., active vertices) increases quickly with respect to the number of edges sampled, whereas with a multi-modal distribution the number of active vertices increases more slowly, leading to denser networks.

\begin{figure}[!htbp]
    \centering
    \includegraphics[scale=.14]{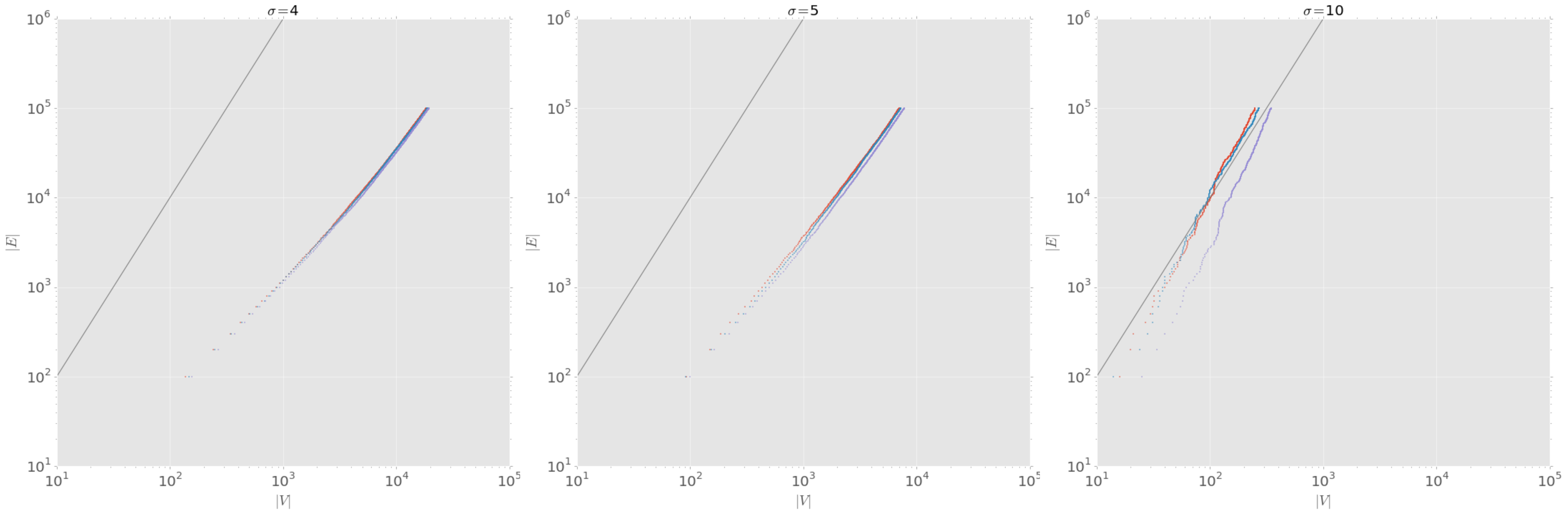}
    \caption{The log-log plots show the number of sampled edges v.s. the number of active vertices of the simulated networks. The $\sigma$ hyper-parameter is set to $4, 5, 10$ (left to right), resulting in $\log |E|/\log |V|$ ratio of approximately $1.5, 1.7, 2.4$ respectively, as shown in the slopes of the colored dots. The different dot colors in each plot represent different random seed. The black solid lines have slopes equal to 2. The scales on the x and y axes are $10^1$ to $10^5$ and $10^6$ respectively.}
    \label{fig:sparsity}
\end{figure}

The quantity of our interest is the ratio of the log number of sampled edges $\log |E|$ to the log number of active vertices $\log |V|$. A network is sparse if the ratio is less than 2 \cite{cai2016edge}. We simulate networks with increasingly more edges at different $\sigma$ values, and show that the sampled networks are sparse for small $\sigma$ in Figure~\ref{fig:sparsity}.

\subsection{Link Predictions}
We conducted 3-fold held-out link prediction experiments using 3 temporal binary network data sets, and benchmarked the proposed model (ATTAS) against the dynamic mixture of mixed-membership stochastic blockmodel (dM${}^3$SB) \cite{ho2011evolving}, aMMSB with Poisson likelihood and two other baselines. We also compared the ATTAS to a variant of the proposed model that does not account for social influence, and instead model the evolution of vertex latent state vectors $\mathbf{h}_v^{(t)}$ with random walk Markov chains. This model is known as RW.
\begin{table}[!htbp]
    \renewcommand{\arraystretch}{1.1}
	\setlength\tabcolsep{1.1pt}
    \caption{3-fold cross-validated mean AUCs with s.e.} \label{tab:auc}
    \begin{tabular}{llll}
        &\bf{ENRON} & \bf{TRADING} & \bf{COLLEGE} \\
        \hline \\
        ATTAS & 0.857$\pm$0.003 & 0.965$\pm$0.001 & 0.823$\pm$0.004 \\
        RW & 0.829$\pm$0.012 & 0.970$\pm$0.001 & 0.811$\pm$0.017\\
        dM${}^3$SB & 0.730$\pm$0.012 & 0.968$\pm$0.002 & 0.656$\pm$0.008\\
        aMMSB & 0.799$\pm$0.006 & 0.731$\pm$0.002 & 0.742$\pm$0.042\\
        Dirich-Mult.& 0.828$\pm$0.006 & 0.946$\pm$0.006 & 0.882$\pm$0.019\\ 
        Equi-prob.& 0.479$\pm$0.005 & 0.553$\pm$0.012 & 0.510$\pm$0.056\\
    \end{tabular}
\end{table}

The performance of the models are compared based on the ROC curves and the AUC metric. In the experiments, the edges observed in the last time slot were randomly split into 3 sets of similar sizes. 2 of the sets were then combined with other edges from previous time slots to form the training data set. The trained models were evaluated on the edges in the held-out fold. The experiments were conducted 3 times, holding out one of the 3 folds every time.

\begin{figure}[!htbp]
    \centering
    \includegraphics[scale=0.3]{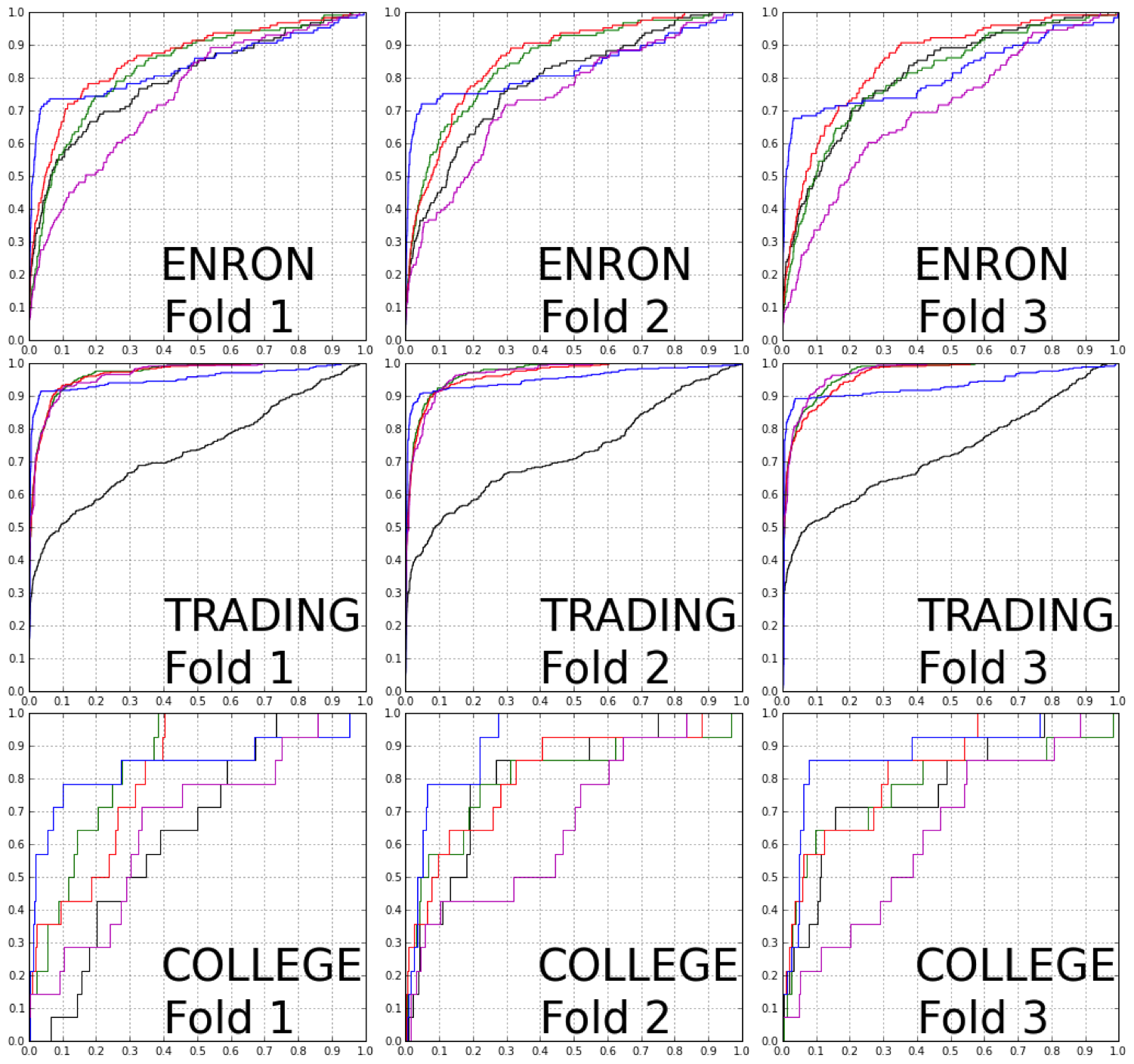}
    \caption{The sub-figures show the ROC curves of ATTAS (red), RW (green), dM${}^3$SB (magenta), aMMSB (black) and Dirichlet-Multinomial (blue) on different folds of the data.}
    \label{fig:roc}
\end{figure}
The results show that the proposed ATTAS and RW models outperformed the baselines in the more complex \textbf{ENRON} and \textbf{TRADING} data sets, while retaining competitive performance in the small \textbf{COLLEGE} data set with less pronounced structures. Most notably, the proposed models outperformed the existing dM${}^3$SB dynamic model in all data sets, achieving better AUCs and higher sensitivities at different false positive rates (fpr) as shown in Figure~\ref{fig:roc}.

While the Dirichlet-Multinomial baseline appears to have high AUC scores, it does not predict non-trivial edges well as shown by its relatively flat ROC curves in Figure~\ref{fig:roc}. For example, in the second fold of the \textbf{ENRON} data, the slope of Dirichlet-Multinomial's ROC curve at 0.2 False Positive Rate (x-axis, approximately where the blue and green/red curves crossed) is near zero while the slopes of ATTAS/RW ROCs are significantly higher. Overall, our models achieved higher sensitivities than Dirichlet-Multinomial at fpr$\geq$0.2 (0.3 for the smallest network) as reported in the ROC curves. In tasks where predicting non-obvious edges (i.e. historically infrequent interactions) are crucial, our models are preferable. In addition, the predictions made by ATTAS are also interpretable because of the built-in community detection mechanism, whereas the Dirichlet-Multinomial baseline only make predictions based on the frequency of interactions between two vertices in the training data.

The results in Table~\ref{tab:auc} also show that modeling social influence is important and highly beneficial for link predictions. The ATTAS model that captures the social influence effect significantly outperformed the RW model with simple random walk state-space in the \textbf{ENRON} and \textbf{COLLEGE} data sets, while achieving essentially the same performance as RW in \textbf{TRADING}.

We describe the data sets used in the link prediction experiments in the following paragraphs. Detailed descriptions of the baseline models are presented in the Supplementary Material because of space constraint.

\textbf{ENRON} \cite{enron} 4 months of ENRON e-mail communication networks were used in the experiments. Two persons/vertices in the networks share an edge in a particular month if there is at least 1 e-mail communication between the pair in that month. There are 126 vertices in the first network, and the number of vertices increased to 138 by the end of the fourth month. We assumed the number of communities to be $3$ in the predictive experiments.

\textbf{TRADING} \cite{trading} 4 years of the international trading networks from 1970 to 1973 were used in the experiments. Two countries/vertices in the networks share an edge in a particular year if the amount of trade between the two countries in the year is non-zero. There are 126 vertices in the network at 1970, and the number increased to 134 by the end of 1973. We assumed the number of communities to be $4$ in the predictive experiments.

\textbf{COLLEGE} \cite{college} This data set consists of 7 snapshots of friendship networks between university freshmen in a Dutch university. The original data set consists of pair-wise friendliness scores between the freshmen surveyed, with score of -1 indicating animosity, to +3 indicating a best friend. We pre-processed the 7 snapshots such that two freshmen/vertices share an edge only if both rated each other with a positive score in the same period. There are 4 vertices in the first network, and the number increased to 31 by the end of the seventh period. We assumed the number of communities to be $3$ in the predictive experiments.

\subsection{Community Detection}
We demonstrate that the proposed ATTAS model can infer meaningful community structure in real temporal networks by fitting the model with $M=2$ to a sequence of 3 temporal networks created from the first 9 months of the 109th US Congress voting records \cite{congress}. The data set was divided into $3$ three-month periods. Two senators share an edge within each of the periods if they casted the same votes for at least $50\%$ of the bills voted on within the period.

\begin{figure}[!htbp]
    \centering
    \includegraphics[scale=0.22]{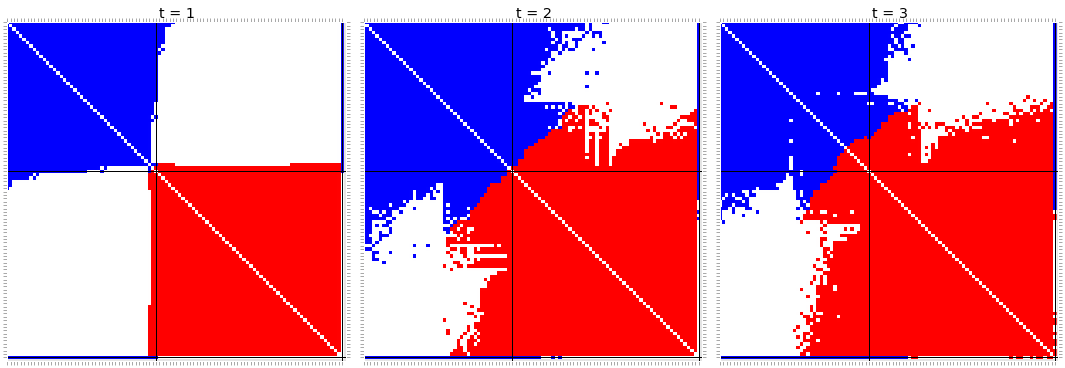}
    \caption{Adjacency matrices of US Congress with the edges colored according to the inferred community types.}
    \label{fig:community}
\end{figure}

Figure~\ref{fig:community} shows the adjacency matrices of the temporal networks. The edges are colored according to their inferred types and the vertices (rows and columns) are sorted according to the senators' party affiliations, with the black lines separating the Democrats (left/top of the black lines) from the Republicans. A single unaffiliated senator is represented in the bottom row/right-most column. Within each party, the senators are sorted according to their relative frequencies of participating in each edge type.

The homogeneous edge colors in the top-left and bottom-right quadrants of the adjacency matrices clearly show that ATTAS was able to infer meaning community structures that reflect the reality (i.e., party affiliations). We also observed that the sizes of the inferred communities changed over time, reflecting the changing positions of senators with different views from their fellow party members (e.g., Democrats with more conservative views) and the dominance of the Republican party during the Bush administration. It is interesting to notice that the voting patterns of some senators aligned well with other senators from both parties. They are highly represented in both communities and are likely to form edges with both Republicans and Democrats, acting as hubs in the networks.

\section{Conclusions}\label{sec:conclude}
We proposed an edge exchangeable model for temporal networks that can model sparsity, community structures and social influences in the networks. The proposed model is also less computationally demanding compared to many existing models. Our experiments show that the proposed model excels in link predictions and can recover meaningful community structures.

\bibliographystyle{plain}
\bibliography{dnetwork}

\begin{thebibliography}{10}

\bibitem{congress}
109th senate roll call data.
\newblock \url{http://www.voteview.com/senate109.htm}.
\newblock Accessed: 2017-04-22.

\bibitem{trading}
Economics web institute world trade.
\newblock \url{http://www.economicswebinstitute.org/worldtrade.htm}.
\newblock Accessed: 2017-05-18.

\bibitem{airoldi2008mixed}
Edoardo~M Airoldi, David~M Blei, Stephen~E Fienberg, and Eric~P Xing.
\newblock Mixed membership stochastic blockmodels.
\newblock {\em Journal of Machine Learning Research}, 9(Sep):1981--2014, 2008.

\bibitem{aldous1981representations}
David~J Aldous.
\newblock Representations for partially exchangeable arrays of random
  variables.
\newblock {\em Journal of Multivariate Analysis}, 11(4):581--598, 1981.

\bibitem{archer2015black}
Evan Archer, Il~Memming Park, Lars Buesing, John Cunningham, and Liam Paninski.
\newblock Black box variational inference for state space models.
\newblock {\em arXiv preprint arXiv:1511.07367}, 2015.

\bibitem{barabasi2016network}
Albert-László Barabási and Márton Pósfai.
\newblock {\em Network science}.
\newblock Cambridge University Press, Cambridge, 2016.

\bibitem{blei2017variational}
David~M Blei, Alp Kucukelbir, and Jon~D McAuliffe.
\newblock {Variational Inference: A Review for Statisticians}.
\newblock {\em Journal of the American Statistical Association},
  112(518):859--877, 2017.

\bibitem{blei2006dynamic}
David~M Blei and John~D Lafferty.
\newblock Dynamic topic models.
\newblock In {\em Proceedings of the 23rd international conference on Machine
  learning}, pages 113--120. ACM, 2006.

\bibitem{blundell2012modelling}
Charles Blundell, Jeff Beck, and Katherine~A Heller.
\newblock Modelling reciprocating relationships with hawkes processes.
\newblock In {\em Advances in Neural Information Processing Systems}, pages
  2600--2608, 2012.

\bibitem{cai2016edge}
Diana Cai, Trevor Campbell, and Tamara Broderick.
\newblock Edge-exchangeable graphs and sparsity.
\newblock In {\em Advances in Neural Information Processing Systems}, pages
  4242--4250, 2016.

\bibitem{caron2014sparse}
François Caron and Emily~B. Fox.
\newblock Sparse graphs using exchangeable random measures.
\newblock {\em Journal of the Royal Statistical Society: Series B (Statistical
  Methodology)}, pages n/a--n/a.

\bibitem{crane2016edge}
Harry Crane and Walter Dempsey.
\newblock Edge exchangeable models for network data.
\newblock {\em arXiv preprint arXiv:1603.04571}, 2016.

\bibitem{farajtabar2015coevolve}
Mehrdad Farajtabar, Yichen Wang, Manuel~Gomez Rodriguez, Shuang Li, Hongyuan
  Zha, and Le~Song.
\newblock Coevolve: A joint point process model for information diffusion and
  network co-evolution.
\newblock In {\em Advances in Neural Information Processing Systems}, pages
  1954--1962, 2015.

\bibitem{ghahramani1997factorial}
Zoubin Ghahramani, Michael~I Jordan, and Padhraic Smyth.
\newblock Factorial hidden markov models.
\newblock {\em Machine learning}, 29(2-3):245--273, 1997.

\bibitem{goldenberg2010survey}
Anna Goldenberg, Alice~X Zheng, Stephen~E Fienberg, Edoardo~M Airoldi, et~al.
\newblock A survey of statistical network models.
\newblock {\em Foundations and Trends{\textregistered} in Machine Learning},
  2(2):129--233, 2010.

\bibitem{gopalan2012scalable}
Prem~K Gopalan, Sean Gerrish, Michael Freedman, David~M Blei, and David~M
  Mimno.
\newblock Scalable inference of overlapping communities.
\newblock In {\em Advances in Neural Information Processing Systems}, pages
  2249--2257, 2012.

\bibitem{heaukulani2013dynamic}
Creighton Heaukulani and Zoubin Ghahramani.
\newblock Dynamic probabilistic models for latent feature propagation in social
  networks.
\newblock In {\em ICML (1)}, pages 275--283, 2013.

\bibitem{ho2011evolving}
Qirong Ho, Le~Song, and Eric~P. Xing.
\newblock Evolving cluster mixed-membership blockmodel for time-evolving
  networks.
\newblock In {\em Proceedings of the Fourteenth International Conference on
  Artificial Intelligence and Statistics, {AISTATS} 2011, Fort Lauderdale, USA,
  April 11-13, 2011}, pages 342--350, 2011.

\bibitem{hoover1979relations}
Douglas~N Hoover.
\newblock Relations on probability spaces and arrays of random variables.
\newblock {\em Preprint, Institute for Advanced Study, Princeton, NJ}, 2, 1979.

\bibitem{ishiguro2010dynamic}
Katsuhiko Ishiguro, Tomoharu Iwata, Naonori Ueda, and Joshua~B Tenenbaum.
\newblock Dynamic infinite relational model for time-varying relational data
  analysis.
\newblock In {\em Advances in Neural Information Processing Systems}, pages
  919--927, 2010.

\bibitem{janson2017edge}
Svante Janson.
\newblock On edge exchangeable random graphs.
\newblock {\em arXiv preprint arXiv:1702.06396}, 2017.

\bibitem{kingma2014adam}
Diederik Kingma and Jimmy Ba.
\newblock Adam: A method for stochastic optimization.
\newblock {\em arXiv preprint arXiv:1412.6980}, 2014.

\bibitem{kingma2013auto}
Diederik~P Kingma and Max Welling.
\newblock Auto-encoding variational bayes.
\newblock {\em arXiv preprint arXiv:1312.6114}, 2013.

\bibitem{linderman2014discovering}
Scott Linderman and Ryan Adams.
\newblock Discovering latent network structure in point process data.
\newblock In {\em International Conference on Machine Learning}, pages
  1413--1421, 2014.

\bibitem{luong2015effective}
Minh-Thang Luong, Hieu Pham, and Christopher~D Manning.
\newblock Effective approaches to attention-based neural machine translation.
\newblock {\em arXiv preprint arXiv:1508.04025}, 2015.

\bibitem{ng2016scaling}
Yin~Cheng Ng, Pawel~M Chilinski, and Ricardo Silva.
\newblock Scaling factorial hidden markov models: Stochastic variational
  inference without messages.
\newblock In {\em Advances in Neural Information Processing Systems}, pages
  4044--4052, 2016.

\bibitem{orbanz2015bayesian}
Peter Orbanz and Daniel~M Roy.
\newblock Bayesian models of graphs, arrays and other exchangeable random
  structures.
\newblock {\em IEEE transactions on pattern analysis and machine intelligence},
  37(2):437--461, 2015.

\bibitem{sarkar2006dynamic}
Purnamrita Sarkar and Andrew~W Moore.
\newblock Dynamic social network analysis using latent space models.
\newblock In {\em Advances in Neural Information Processing Systems}, pages
  1145--1152, 2006.

\bibitem{saul1996exploiting}
Lawrence~K Saul and Michael~I Jordan.
\newblock Exploiting tractable substructures in intractable networks.
\newblock In {\em Advances in neural information processing systems}, pages
  486--492, 1996.

\bibitem{sewell2017latent}
Daniel~K Sewell, Yuguo Chen, et~al.
\newblock Latent space approaches to community detection in dynamic networks.
\newblock {\em Bayesian Analysis}, 12(2):351--377, 2017.

\bibitem{enron}
Jitesh Shetty and Jafar Adibi.
\newblock The enron email dataset database schema and brief statistical report.
\newblock {\em Information sciences institute technical report, University of
  Southern California}, 4, 2004.

\bibitem{college}
Gerhard~G. Van~de Bunt, Marijtje A.~J. Van~Duijn, and Tom A.~B. Snijders.
\newblock Friendship networks through time: An actor-oriented dynamic
  statistical network model.
\newblock {\em Computational and Mathematical Organization Theory},
  5(2):167--192, 1999.

\bibitem{williamson2016nonparametric}
Sinead~A Williamson.
\newblock Nonparametric network models for link prediction.
\newblock {\em Journal of Machine Learning Research}, 17(202):1--21, 2016.

\bibitem{xing2010state}
Eric~P Xing, Wenjie Fu, Le~Song, et~al.
\newblock A state-space mixed membership blockmodel for dynamic network
  tomography.
\newblock {\em The Annals of Applied Statistics}, 4(2):535--566, 2010.

\bibitem{xu2014dynamic}
Kevin~S Xu and Alfred~O Hero.
\newblock Dynamic stochastic blockmodels for time-evolving social networks.
\newblock {\em IEEE Journal of Selected Topics in Signal Processing},
  8(4):552--562, 2014.

\bibitem{zreik2017dynamic}
Rawya Zreik, Pierre Latouche, and Charles Bouveyron.
\newblock The dynamic random subgraph model for the clustering of evolving
  networks.
\newblock {\em Comput. Stat.}, 32(2):501--533, June 2017.

\end{thebibliography}
\end{document}


\onecolumn

\aistatstitle{Supplementary Material: A Dynamic Edge Exchangeable Model for Sparse Temporal Networks}

\section{Variational Inference}\label{sec:vi}
We derive the ELBO and the relevant update equations for the proposed dynamic network model in this section. The derived algorithm was implemented using TensorFlow \cite{abadi2016tensorflow}.

\subsection{Model Joint Distribution}\label{sec:joint}
We derive the proposed model's joint distribution for a temporal network data set $\{\Gt\}_{t=1}^T$ and assume $M$ communities in the model.

Each undirected network $\Gt = (\Vt, \Et)$ consists of $|\Vt|=Q^{(t)}$ vertices $\Vt = \{1, \ldots, Q^{(t)}\}$ and $|\Et| = N^{(t)}$ edges $\Et = \{e_1^{(t)}, \ldots, e_{N^{(t)}}^{(t)}\}$ where $\eit = (\vit, \vpit)$ and $\vit, \vpit \in \Vt$. New vertices are added to the vertex set at each time step as they participate in at least one observed edge. Therefore, $V^{(t-1)} \subset \Vt$ and $V^{(0)} = \{\}$.

We introduce the following latent variables to model the network temporal dependency, and the dependency within each network.
\begin{itemize}
    \item $\cit$: edge community type indicator for edge $\eit$.
    \item $\kt$: $\mathbf{R}^{M \times 1}$ latent vector parameterizing the multivariate logistic normal distributions where $\cit$ are sampled.
    \item $\mathbf{h}_v^{(t)}$: $\mathbf{R}^{M \times 1}$ latent state vector for vertex $v$ first observed at $\tau_v$. The $t$ index enumerates from $\tau_v$ to $T$.
    \item $\mathbf{h}_z^{(t)}$: $\mathbf{R}^{M \times 1}$ latent state vector for unobserved potential new vertex $z$. We denote the set of unobserved potential new vertices at $t$ as $V_z^{(t)}$
    \item $L^{(t)}$: The sum of the number of unobserved potential new vertices and the newly observed vertices at $t$.
    \item $\lambda^{(t)}$: The Poisson log-rate parameter for $L^{(t)}$.
\end{itemize}

The joint probability distribution of the observed temporal networks and latent variables conditioning on model parameters $\theta = \{\mu_\lambda$, $\sigma_\lambda$, $a_\lambda$, $\bm{\mu}$, $\mathbf{B}$, $\bm{\mu}_k$, $\mathbf{A}_k$, $\mathbf{B}_k\}$ is as follow.

\begin{align*}\label{eqn:model_joint}
    &p_\theta(\{\Gt\}_{t=1}^T, \{\{\cit\}_{i=1}^{N^{(t)}}\}_{t=1}^{T}, \{\mathbf{h}_v^{(\tau_v:T)}\}_{v=1}^{Q^{(T)}}, \{\mathbf{h}_z^{(t)}|z \in V_z^{(t)}\}_{t=1}^{T}, \{L^{(t)}\}_{t=1}^T, \{\lambda^{(t)}\}_{t=1}^T) \numberthis\\
    =&(\prod_{t=1}^{T}\prod_{i=1}^{N^{(t)}}\prod_{j\in\eit}P_t(v = j|\cit, \{\mathbf{h}_v^{(t)}|v \in \Vt\}, \{\mathbf{h}_z^{(t)}|z \in V_z^{(t)}\})P_t(c = \cit|\kt))\\
    &(\prod_{t=1}^{T}\prod_{z \in V_z^{(t)}}p_{\theta}(\mathbf{h}_z^{(t)}|\bm{\mu}, \mathbf{B}))(\prod_{v \in V^{(T)}}p_{\theta}(\mathbf{h}_v^{(\tau_v)})\prod_{t=\tau_v+1}^{T}p_{\theta}(\mathbf{h}_v^{(t+1)}|\Gt, \{\mathbf{h}_i^{(t)}|i\in\Vt\}))\\
    &(p_\theta(\mathbf{k}^{(1)})\prod_{t=2}^Tp_\theta(\kt|\mathbf{k}^{(t-1)}))(P(L^{(1)}|\lambda^{(1)})p_\theta(\lambda^{(1)})\prod_{t=2}^{T}P(L^{(t)}|\lambda^{(t)})p_\theta(\lambda^{(t)}|\lambda^{(t-1)}))
\end{align*}
The probability distributions specified in Equation~\ref{eqn:model_joint} are detailed in Section 2 of the main text.

\subsection{Variational Distributions}\label{sec:qdist}
We introduce the approximating variational distribution $q_\beta$ with a set of variational parameters $\beta$ for the latent variables introduced in Section~\ref{sec:joint}. The variational distributions belong to the structured mean-field exponential family, and preserve the time dependency of the latent variables.
\begin{itemize}
    \item $q_\beta(\cit)$: Categorical distribution with M categories and parameters $\bm{\pi}_i^{(t)}$
    \item $q_\beta(\mathbf{k}^{(1:T)})$: $M$ dimensional Gaussian Markov chain. We assumed the Markov chain factorizes across the $M$ dimensions to reduce the number of variational parameters.
    \item $q_\beta(\mathbf{h}_v^{(\tau_v:T)})$: $M$ dimensional Gaussian Markov chain. We assumed the Markov chain factorizes across the $M$ dimensions to reduce the number of variational parameters.
    \item $q_\beta(\mathbf{h}_z^{(t)})$: This is the prior distribution $p_{\theta}(\mathbf{h}_v^{(\tau_v)})$ as the vertices in $V_z^{(t)}$ are never observed.
    \item $q_\beta(L^{(t)})$: Shifted Poisson Distribution with rate variational parameter $\eta^{(t)}$. The Poisson distribution is shifted to the right by $|\Vt| - |V^{(t-1)}|$ because  $L^{(t)} \geq |\Vt| - |V^{(t-1)}|$ a posteriori.
    \item $q_\beta(\lambda^{(1:T)})$: $1$ dimensional Gaussian Markov chain.
\end{itemize}

\subsection{Evidence Lower Bound (ELBO)}
The ELBO $\ELBO$ is the variational objective function for estimating the variational parameters $\beta$ and learning model parameters $\theta$. ELBO is the sum of the expectation of log model joint distribution in Equation~\ref{eqn:model_joint} with respect to $q_\beta$ and the entropy of $q_\beta$.

\begin{align*}\label{eqn:elbo}
    \ELBO &= \sum_{t=1}^T\sum_{i=1}^{N^{(t)}}\Eq[\ln P_t(\cit|\kt)]+\sum_{j \in \eit}\Eq[\ln P_t(v = j|\cit, \{\mathbf{h}_v^{(t)}|v \in \Vt\}, \{\mathbf{h}_z^{(t)}|z \in V_z^{(t)}\})] \numberthis\\
    & +\sum_{v \in V^{(T)}}\{\Eq[\ln p_{\theta}(\mathbf{h}_v^{(\tau_v)})]+\sum_{t=\tau_v+1}^{T}\Eq[\ln p_{\theta}(\mathbf{h}_v^{(t+1)}|\Gt, \{\mathbf{h}_i^{(t)}|i\in\Vt\})]\}\\
    & + \Eq[\ln p_\theta(\mathbf{k}^{(1)})] + \sum_{t=2}^T\Eq[\ln p_\theta(\kt|\mathbf{k}^{(t-1)})] + \Eq[\ln p_\theta(\lambda^{(1)})] + \sum_{t=2}^{T}\Eq[\ln p_\theta(\lambda^{(t)}|\lambda^{(t-1)})]\\
    & + \sum_{t=1}^{T}\{\Eq[\ln P(L^{(t)}|\lambda^{(t)})] + \eta^{(t)}\Eq[\ln p_{\theta}(\mathbf{h}_z^{(t)}|\bm{\mu}, \mathbf{B})]\} - \Eq[\ln q_\beta]
\end{align*}

The $\eta^{(t)}\Eq[\ln p_{\theta}(\mathbf{h}_z^{(t)}|\bm{\mu}, \mathbf{B})]$ term in Equation~\ref{eqn:elbo} is contributed by the product of prior distributions for the latent vectors $\mathbf{h}_z^{(t)}$ of unobserved potential new vertices. As these vertices are never observed, their numbers are uncertain and the uncertainty is accounted for in $q_\beta(L^{(t)})$. Therefore, the expected numbers of unobserved potential new vertices are $\eta^{(1:T)}$. As we explained in Section~\ref{sec:qdist}, the variational distribution $q_\beta(L^{(t)})$ is set as the prior distribution $p_{\theta}(\mathbf{h}_z^{(t)}|\bm{\mu}, \mathbf{B})$. The factor $\eta^{(t)}\Eq[\ln p_{\theta}(\mathbf{h}_z^{(t)}|\bm{\mu}, \mathbf{B})]$ is a multiple of the negative entropy of $p_{\theta}(\mathbf{h}_z^{(t)}|\bm{\mu}, \mathbf{B})$ and cancels out the corresponding entropy term in $\Eq[\ln q_\beta]$.

\subsection{Bounding the Logistic Normal Distributions}\label{sec:bound}
Computing the expected log-normalizing constants of the logistic normal distributions $P_t(\cit|\kt)$ and $P_t(v|\cit, \{\mathbf{h}_v^{(t)}|v \in \Vt\}, \{\mathbf{h}_z^{(t)}|z \in V_z^{(t)}\})$ is intractable as $-\Eq[\ln (\sum_{m=1}^{M} e^{k_m^{(t)}})]$ and $-\Eq[\ln(\sum_{v\in\Vt}e^{h_{v,m}^{(t)}}+ \sum_{z \in V_z^{(t)}}e^{h_{z,m}^{(t)}})]$ cannot be analytically evaluated. We apply the bound $-\ln Z \geq -\frac{Z}{\zeta} - \ln \zeta + 1$ to linearize the log-sum-exp expression such that their expected linear approximations can be evaluated analytically \cite{blei2006dynamic}. The additional parameters $\zeta$ introduced by the bound are additional variational parameters that can be optimized.

Applying the bound to $\ln (\sum_{m=1}^{M} e^{k_m^{(t)}})$ and $\ln(\sum_{v\in\Vt}e^{h_{v,m}^{(t)}}+ \sum_{z \in V_z^{(t)}}e^{h_{z,m}^{(t)}})$ yields the following expressions
\begin{equation}
	- \Eq[\ln{(\sum_{v\in\Vt}e^{h_{v,m}^{(t)}}+\sum_{z\in V_z^{(t)}}e^{h_{z,m}^{(t)}})}] \geq - \frac{1}{\zeta_{m}^{(t)}}(\sum_{v\in\Vt}\Eq[e^{h_{v,m}^{(t)}}]+\eta^{(t)}\Eq[e^{h_{z,m}^{(t)}}]) - \ln \zeta_{m}^{(t)} + 1
	\label{eqn:bound1}
\end{equation}

\begin{equation}
	- \Eq[\ln{(\sum_{m=1}^{M} e^{k_m^{(t)}})}] \geq - \frac{1}{\zeta_c^{(t)}}\sum_{m=1}^{M} \Eq[e^{k_m^{(t)}}] - \ln \zeta_c^{(t)} + 1.
	\label{eqn:bound2}
\end{equation}
The expectations of the bounds can be evaluated analytically as $\Eq[e^{h_{v,m}^{(t)}}]$, $\Eq[e^{h_{z,m}^{(t)}}]$ and $\Eq[e^{k_{m}^{(t)}}]$ are simply expectations of log-normal random variables. 

The fixed point update equations for the $T+MT$ variational parameters $\{\{\zeta_m^{(t)}\}_{m=1}^M\}_{t=1}^T$ and $\{\zeta_c^{(t)}\}_{t=1}^{T}$ can be derived by setting the derivatives of the bounds with respect to the parameters to 0.

\begin{equation}\label{eqn:update1}
    \zeta_m^{(t)} = \sum_{v\in\Vt}\Eq[e^{h_{v,m}^{(t)}}]+\eta^{(t)}\Eq[e^{h_{z,m}^{(t)}}]
\end{equation}

\begin{equation}\label{eqn:update2}
    \zeta_c^{(t)} = \sum_{m=1}^{M} \Eq[e^{k_m^{(t)}}]
\end{equation}

The bounds are tight when the variational parameters are updated according to Equation~\ref{eqn:update1} and \ref{eqn:update2}.

\subsection{Monte Carlo Approximations for ATTAS}
Another source of intractability in ELBO is the expectations of the ATTAS conditional distributions $\Eq[\ln p_{\theta}(\mathbf{h}_v^{(t+1)}|\Gt, \{\mathbf{h}_i^{(t)}|i\in\Vt\})]$. The expectations of the log conditional Gaussian distributions cannot be evaluated analytically because the conditional means are parameterized as non-linear functions of $\{\mathbf{h}_i^{(t)}|i\in\Vt\}$ as described in Section 2 of the main text.

Fortunately, the Gaussian $q_\beta(\mathbf{h}_v^{(\tau_v:T)})$ allows the expectations to be approximated stochastically using unbiased Monte Carlo samples. The gradients of the parameters in $q_\beta(\mathbf{h}_v^{(\tau_v:T)})$ can also be approximated stochastically with unbiased Monte Carlo samples. The stochastic gradients computed using only a single Monte Carlo sample and the reparameterization tricks \cite{kingma2013auto} worked well in our experiments.

\subsection{Update Equations for Edge Type Indicators $\cit$}
The variational parameters of $q_\beta(\cit)$, $\bm{\pi}_i^{(t)} = [\pi_{i,1}^{(t)}, \ldots, \pi_{i,M}^{(t)}]$ where $\sum_{m=1}^M\pi_{i,m}^{(t)}$, can be updated analytically by exploiting the conjugate structures in the relevant parts of $\ELBO$ and the linear log bounds in Equation~\ref{eqn:bound1} and \ref{eqn:bound2}.

Isolating the relevant terms in $\ELBO$ and applying the log-bounds to linearize the expected log-normalizing constants, 
\begin{align*}\label{eqn:elbo_pi}
    \hat{\mathcal{L}}(\bm{\pi}_i^{(t)}) = &\sum_{m=1}^M\pi_{i,m}^{(t)}\{\Eq[h_{\vit,m}^{(t)}]+\Eq[h_{\vpit,m}^{(t)}]-\frac{2}{\zeta_m^{(t)}}(\sum_{v\in\Vt}\Eq[e^{h_{v,m}^{(t)}}]+\eta^{(t)}\Eq[e^{h_{z,m}^{(t)}}]) - 2\ln \zeta_m^{(t)} + 2\} \numberthis\\
    &+\sum_{m=1}^M\pi_{i,m}^{(t)}\{\Eq[k_m^{(t)}] - \frac{1}{\zeta_c^{(t)}}\sum_{m=1}^{M} \Eq[e^{k_m^{(t)}}] - \ln \zeta_c^{(t)} + 1\}- \sum_{m=1}^M\pi_{i,m}^{(t)} \ln \pi_{i,m}^{(t)}
\end{align*}

Setting the derivative of Equation~\ref{eqn:elbo_pi} with respect to $\pi_{i,m}^{(t)}$ to 0 results in the coordinate descent update equation
\begin{align*}\label{eqn:elbo_pi_update}
    \ln \pi_{i,m}^{(t)} =& \Eq[h_{\vit,m}^{(t)}]+\Eq[h_{\vpit,m}^{(t)}]-\frac{2}{\zeta_m^{(t)}}(\sum_{v\in\Vt}\Eq[e^{h_{v,m}^{(t)}}]+\eta^{(t)}\Eq[e^{h_{z,m}^{(t)}}]) - 2\ln \zeta_m^{(t)} + 2 \numberthis\\
    &+ \Eq[k_m^{(t)}] - \frac{1}{\zeta_c^{(t)}}\sum_{m=1}^{M} \Eq[e^{k_m^{(t)}}] - \ln \zeta_c^{(t)}
\end{align*}

The sums over the vertices in $\Vt$ in Equation~\ref{eqn:elbo_pi_update} and \ref{eqn:update1} are the computational bottleneck of the variational inference algorithm, as they scale linearly with respect to the number of observed vertices. Fortunately, each summation only has to be computed once per iteration and applies to all $\cit$.

\section{Simulation Experiment}
We simulated a temporal network with 3 time steps and 2 communities using the dynamic edge exchangeable network model with the RW state-space. Parallel edges in the simulated networks were collapsed to 1, with the ground truth edge type indicator random variable assigned through majority voting. Ties were broken by random selections.

\begin{figure}[!htbp]
    \centering
    \includegraphics[scale=0.45]{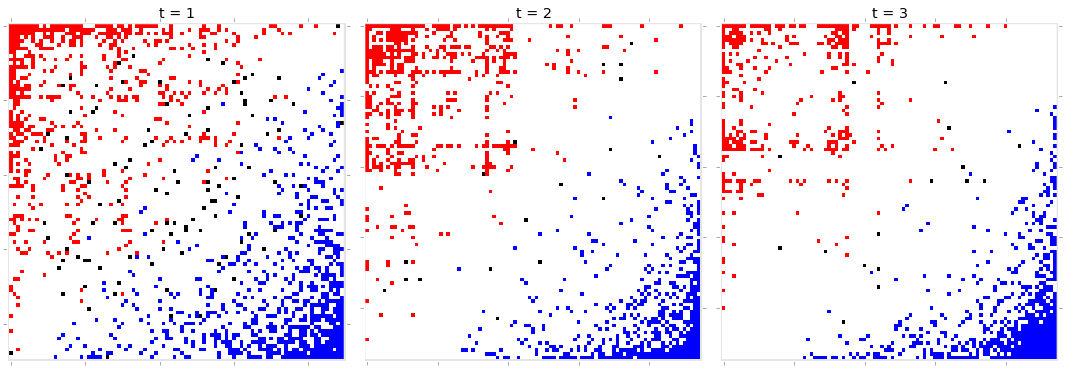}
    \caption{Adjacency matrices of the simulated temporal network. The red and blue edges correspond to the two classes of edges that are correctly classified with the variational algorithm. The black edges are the mis-classified edges.}
    \label{fig:sim_adj}
\end{figure}

The variational inference algorithm recovered 96$\%$ of the ground truth edge type indicators $\cit$ at a normalized mutual information (NMI) score of 0.75 using the variational inference algorithm. The adjacency matrices of the simulated network is shown in Figure~\ref{fig:sim_adj}.

\section{Link Prediction Experiment Model Descriptions}
The following paragraphs describe the details of the models compared in the link prediction 3-fold cross-validation experiment.

\textbf{ATTAS, RW} The proposed dynamic network model with the ATTAS/random walk state-space. The models were trained for $50,000$ iterations and given $5$ random restarts per experiment. The predictive probability of seeing an edge between vertex $i$ and $j$ were computed using $500$ Monte Carlo samples drawn from the fitted variational distributions.

\textbf{dM${}^3$SB} The dynamic mixture of mixed-membership stochastic blockmodel proposed in \cite{ho2011evolving}. The model hyper-parameters were selected using the BIC grid search procedure proposed in \cite{ho2011evolving}. The hyper-parameter grids for the number of mixture component and the number of community are $[2, 3, 4, 5]$ and $[3, 4, 5, 6]$ respectively. We performed 5 random restarts per configuration. The model was also modified to leave out the links in the hold-out set.

\textbf{aMMSB} This is the assortative MMSB proposed in \cite{gopalan2012scalable} with Poisson likelihood. All the edges observed in the training data set were aggregated and modelled as counts. The models were trained to convergence and given $5$ random restarts per experiment. The predictive probability is the probability of observing at least one edge between two vertices conditioning on the training data.

\textbf{Dirichlet-Mult.} The Dirichlet-multinomial distributions over edges is equivalent to an Infinite Relational Model \cite{kemp2006learning} where each pair of vertices is in its own cluster. Please refer to \cite{williamson2016nonparametric} for details.

\textbf{Equi-probable} Equi-probable links baseline \cite{williamson2016nonparametric}. This baseline assumes the probability of observing an edge between two vertices is $\frac{1}{N \times (N-1)}$, where $N$ is the number of vertices in the training data.

\bibliographystyle{plain}
\bibliography{dnetwork}